\documentclass[runningheads,a4paper]{llncs}

\usepackage{graphicx}
\usepackage{booktabs}
\usepackage{amsmath,amssymb}
\usepackage{url}
\usepackage{microtype}
\usepackage{float}
\usepackage{placeins}
\begin{document}
\title{Signal Simplification Is Not Predictive Simplification: Diagnosing Residual Neural Forecasting in Short-Horizon Volatility}
\titlerunning{Signal Simplification in Residual Neural Forecasting}

\author{
Bingqi Lian \and
Linfeng Cheng \and
Mei Lu \and
Jerry Wu
}

\authorrunning{B. Lian et al.}

\institute{
Science Academy, College of Computer, Mathematical, and Natural Sciences,
University of Maryland, College Park, MD 20742, USA
\\
\email{\{blian12,lcheng13,meilu\}@umd.edu}
\and
Department of Electrical and Computer Engineering,
University of Maryland, College Park, MD 20742, USA
\\
\email{jerrywu@umd.edu}
}
\maketitle

\begin{abstract}
Hybrid statistical--neural pipelines often assume that a successful statistical first stage leaves a cleaner and more learnable residual target. We examine that assumption in short-horizon volatility forecasting through a Signal--Forecast--System diagnostic framework. Across five liquid U.S. assets, a volatility-aligned HAR-style model outperforms AR, MA, and ARIMA. Within expanding training windows, the pre-standardization fitted residual process used to construct residual-LSTM sequences has about 82\% lower variance than the corresponding target and near-zero lag-1 autocorrelation; independently, rolling pseudo-out-of-sample HAR errors show about 74\% variance reduction and similarly weak lag-1 dependence. Residual-only LSTM augmentation nevertheless raises mean squared error from 0.3049 to 0.3594 on average, with deterioration on every asset. Pure LSTM records the lowest selected pseudo-out-of-sample MSE, 0.2649, while the residual hybrid requires substantially more end-to-end runtime without improving accuracy. We describe this pattern as forecaster--preconditioner asymmetry: first-stage forecasting success and statistical residual simplification need not translate into useful downstream neural preconditioning.
\end{abstract}

\section{Introduction}
Short-horizon volatility forecasts support intraday risk monitoring, portfolio adjustment, margining, and trading infrastructure. Classical realized-volatility research shows that volatility is strongly persistent and that heterogeneous horizons are useful for forecasting its dynamics \cite{andersen2003modeling,corsi2009simple}. Machine-learning studies likewise find that persistence remains a central source of predictive information in realized-variance forecasting \cite{bucci2020realized,christensen2023machine}. In this study, volatility is represented by a six-observation rolling variance proxy constructed from squared close-to-close returns between consecutive regular-session bars. The target is persistent and noisy, with short-horizon dynamics that may vary over time, making it a useful setting for studying how statistical persistence modeling interacts with neural sequence learning.

Hybrid statistical--neural forecasting models are commonly motivated by a decomposition argument. A statistical model first captures structured temporal dependence, after which a neural model learns the remaining residual component. This logic underlies ARIMA--ANN and related residual-learning hybrids \cite{zhang2003time,khashei2011novel} and has also motivated combinations of neural networks with conditional-volatility models in finance \cite{kim2018forecasting}. Recent volatility evidence is more nuanced: nonlinear machine-learning models do not uniformly dominate linear or HAR-family alternatives, and gains from machine learning can depend on the information set supplied to the model \cite{branco2024forecasting,luo2024forecasting,gunnarsson2024prediction}. Residual compression, then, is only part of the hybrid-design problem. The transformed series must still retain structure that the downstream learner can use.

That observation separates two roles of the same first-stage model. As a \emph{forecaster}, it is judged by the quality of its own predictions; as a \emph{preconditioner}, it is judged by the usefulness of the representation passed to the next learner. The two roles need not move together. A strong statistical forecast may absorb persistent structure that a neural sequence model could otherwise exploit, leaving a residual that is smaller in variance but no easier to predict. We refer to this empirical pattern as \emph{forecaster--preconditioner asymmetry}.

We examine this possibility using a focused six-model design with distinct diagnostic roles. AR, MA, and ARIMA provide generic statistical references. An observation-scale HAR-style specification provides a volatility-aligned statistical first stage. A pure LSTM serves as the raw-sequence neural control, while HAR-style+LSTM tests whether residual-only neural augmentation adds value after the stronger statistical model. Forecasting performance is evaluated chronologically, consistent with rolling-origin practice in time-series evaluation \cite{tashman2000out,hewamalage2023forecast}.

The same pattern appears across the three diagnostic layers. HAR-style is substantially stronger than the generic statistical references. The pre-standardization fitted residual process used to construct residual-LSTM training sequences is strongly compressed within expanding windows, with about 82\% lower variance and near-zero lag-1 autocorrelation on average; rolling pseudo-out-of-sample HAR errors independently show about 74\% variance reduction and the same qualitative attenuation of immediate dependence. HAR-style+LSTM, however, increases mean squared error from 0.3049 to 0.3594, and the increase appears on all five assets. Pure LSTM records the lowest selected pseudo-out-of-sample MSE, 0.2649, while the residual hybrid also carries substantially higher end-to-end runtime than HAR-style.

Three research questions organize the analysis. \textbf{RQ1:} Does a volatility-aligned statistical first stage provide a credible standalone forecast while materially simplifying the residual signal? \textbf{RQ2:} Does this statistical simplification translate into improved residual-only neural forecasting? \textbf{RQ3:} Does residual neural augmentation improve the accuracy--efficiency trade-off relative to the standalone statistical and neural alternatives?

The contribution is a diagnostic view of residual preconditioning that separates signal transformation, downstream predictive value, and system cost. We evaluate signal transformation for the pre-standardization fitted residual process used to construct neural training sequences and, separately, for rolling forecast errors, avoiding the assumption that these are the same residual object. In short-horizon volatility, that view reveals a forecaster--preconditioner asymmetry: HAR-style forecasts well and materially changes both residual processes, yet the residual-only neural stage worsens predictive performance. The system evidence points the same way, with higher computational cost and lower accuracy than both HAR-style and Pure LSTM.

\section{Related Work}
\subsection{Volatility Persistence and HAR-Style Forecasting}
Realized variance aggregates high-frequency squared returns over a fixed interval and provides an empirical proxy for latent return variation \cite{andersen2003modeling}. High-frequency measures can be affected by market microstructure noise \cite{hansen2006realized}, while volatility itself exhibits clustering and persistent conditional heteroskedasticity \cite{engle1982autoregressive,bollerslev1986generalized}. The HAR-RV model introduced by Corsi captures long-memory-like volatility behavior parsimoniously through components measured at heterogeneous horizons \cite{corsi2009simple}. Its multi-horizon structure remains a standard benchmark in modern volatility forecasting, including comparisons with machine-learning methods \cite{christensen2023machine,branco2024forecasting}. Recent evidence also emphasizes that richer nonlinear models need not systematically dominate simpler linear volatility models \cite{branco2024forecasting}.

Our HAR-style specification adapts this heterogeneous-horizon structure to the ordered intraday-bar setting. Current-, three-observation-, and six-observation components align the HAR idea with the rolling target used in this study, while preserving the multi-scale representation that motivates the original HAR-RV model.

\subsection{Neural and Residual Hybrid Forecasting}
LSTM networks were designed to learn long-range sequential dependencies \cite{hochreiter1997long} and are now widely used in financial time-series forecasting \cite{sezer2020financial,lim2021timeseries}. Neural models have also been evaluated directly for realized-volatility forecasting \cite{bucci2020realized}. More expressive sequence architectures such as temporal convolutional networks, N-BEATS, and transformer-based models broaden the available design space \cite{bai2018empirical,oreshkin2020nbeats,nie2023time}.

Hybrid forecasting models combine statistical and neural components in an attempt to separate structured and nonlinear dynamics. Zhang and Khashei and Bijari use ARIMA-based first stages followed by neural residual learning \cite{zhang2003time,khashei2011novel}. In financial volatility, Kim and Won combine LSTM with GARCH-family models \cite{kim2018forecasting}. More recent work extends HAR-style volatility forecasting with machine-learning information channels and shows that augmentation can be valuable when it contributes additional predictors rather than only a residual target \cite{luo2024forecasting}. A recent review likewise finds that both modern AI methods and traditional econometric models remain relevant in volatility forecasting and identifies hybrid designs as an active research direction \cite{gunnarsson2024prediction}. These studies motivate hybridization, but final hybrid accuracy does not reveal whether residualization itself creates a more learnable target. We use a residual-only design to measure the predictive content retained in the transformed residual separately from additional contextual inputs.

\subsection{Forecast Evaluation}
Forecast evaluation must respect temporal ordering and distinguish model selection from final performance estimation. Rolling-origin evaluation provides repeated chronological forecast origins and is preferable to random partitioning for many dependent time-series settings \cite{tashman2000out,hewamalage2023forecast}. We report MSE, MAE, benchmark-relative out-of-sample $R^2$, and QLIKE. QLIKE is widely used in volatility comparison because it remains useful when volatility is measured with imperfect proxies \cite{patton2011volatility}. We additionally use Diebold--Mariano (DM) comparisons \cite{diebold1995comparing} with Newey--West heteroskedasticity- and autocorrelation-consistent variance estimation \cite{newey1987simple}.

\section{Methodology: Signal--Forecast--System Diagnostic Framework}
\subsection{Preconditioning as a Forecasting Intervention}
Let $y_t$ denote the log six-observation rolling variance proxy at observed bar $t$.
For a statistical first-stage model $m$, two residual objects must be distinguished.
At refitting origin $k$, residual-LSTM training sequences use in-sample fitted residuals
within the available estimation window $\mathcal{T}_k$. Here, ``fit'' denotes the
in-sample fitted values produced by model $m$ after its parameters are estimated on
$\mathcal{T}_k$. Deployment instead produces realized one-step-ahead errors:
\begin{equation}
\tilde e_{t;k}^{(m),\mathrm{fit}}
=
y_t-\hat y_{t;k}^{(m),\mathrm{fit}},
\quad t\in\mathcal{T}_k;
\qquad
e_{t+1}^{(m),\mathrm{POOS}}
=
y_{t+1}-\hat y_{t+1|t}^{(m)}.
\end{equation}
A pseudo-out-of-sample residual enters the history only after its target is observed. A residual-only LSTM estimates a correction $\hat e_{t+1}^{(m)}$, producing
\begin{equation}
 \hat y_{t+1}^{(m+\mathrm{LSTM})}=\hat y_{t+1}^{(m)}+\hat e_{t+1}^{(m)}.
\end{equation}
The hybrid forecast error is therefore
\begin{equation}
 y_{t+1}-\hat y_{t+1}^{(m+\mathrm{LSTM})}=e_{t+1}^{(m),\mathrm{POOS}}-\hat e_{t+1}^{(m)}.
\end{equation}
Thus, residual learning adds value only when the downstream model can forecast the residual better than leaving the first-stage forecast uncorrected. Lower residual variance is not sufficient.

\subsection{Forecaster--Preconditioner Asymmetry}
Conceptually, write the residual as
\begin{equation}
 e_{t+1}^{(m)}=g^{(m)}(z_t)+u_{t+1},
\end{equation}
where $g^{(m)}(z_t)$ is the predictable component left after first-stage forecasting and $u_{t+1}$ is unpredictable variation. A conceptual residual signal-to-noise ratio is
\begin{equation}
 \mathrm{SNR}^{(m)}=\frac{\mathrm{Var}[g^{(m)}(z_t)]}{\mathrm{Var}(u_{t+1})}.
\end{equation}
A successful first-stage model can reduce total residual variance while also reducing the variance of the predictable component. The residual can therefore become statistically smaller without becoming easier to forecast. We use \emph{forecaster--preconditioner asymmetry} for the empirical configuration in which a model forecasts well and strongly simplifies its residual, yet the transformed residual does not support beneficial downstream learning.

\subsection{Signal, Forecast, and System Diagnostics}
The signal diagnostic measures residual variance reduction,
\begin{equation}
 S_m=1-\frac{\mathrm{Var}(e^{(m)})}{\mathrm{Var}(y)},
\end{equation}
and first-order residual autocorrelation. We also report the relative reduction in the magnitude of lag-1 autocorrelation,
\[
 A_m^{\mathrm{ACF}}=1-\frac{|\rho_1(e^{(m)})|}{|\rho_1(y)|}.
\]
This normalized diagnostic equals zero when the residual preserves the raw lag-1 autocorrelation magnitude and approaches one as the residual ACF(1) approaches zero. Together, these quantities characterize how strongly the first stage compresses short-lag statistical structure. We apply them to two distinct residual objects in Table~\ref{tab:signal}. First, for each expanding refit window, we compute variance reduction and ACF diagnostics on the fitted residual history $\tilde e_{t;k}^{(m),\mathrm{fit}}$ actually used to construct residual-LSTM training sequences. Diagnostics are computed within each window before averaging across windows within an asset, preventing older observations from receiving repeated weight merely because they recur in later expanding windows. Second, we compute the same core diagnostics on the sequential HAR-style pseudo-out-of-sample forecast errors $e_{t+1}^{(m),\mathrm{POOS}}$. The two panels therefore characterize the neural training representation and the first-stage rolling forecast errors separately rather than treating them as interchangeable.

The forecast diagnostic uses MSE, MAE, benchmark-relative $R^2$, and QLIKE. Let $\hat\sigma_t^2=\exp(\hat y_t)$ and $\sigma_t^2=\exp(y_t)$. Pointwise QLIKE is
\begin{equation}
 L_t^{\mathrm{QLIKE}}=\log(\hat\sigma_t^2)+\frac{\sigma_t^2}{\hat\sigma_t^2}
 =\hat y_t+\exp(y_t-\hat y_t).
\end{equation}
Lower values are preferred \cite{patton2011volatility}. This form differs from the canonical ratio-based QLIKE expression only by an observation-specific term common to all competing forecasts, so forecast rankings and pairwise loss differentials are unchanged. Benchmark-relative out-of-sample $R^2$ is computed against the one-step persistence forecast $\hat y^{\mathrm{pers}}_{t+1}=y_t$ as
\begin{equation}
R^2_{\mathrm{OOS}}=1-\frac{\sum_t (y_{t+1}-\hat y_{t+1})^2}{\sum_t (y_{t+1}-y_t)^2}.
\end{equation}
For paired forecast comparison, $d_t=L_{a,t}-L_{b,t}$, so positive values indicate larger loss for model $a$. DM statistics use a Newey--West long-run variance estimator with lag 6 \cite{diebold1995comparing,newey1987simple}, covering the six-observation scale highlighted by the residual diagnostics; lag 12 provides a wider-bandwidth robustness check.

The system diagnostic compares predictive loss $L(f)$ and end-to-end wall-clock cost $C(f)$. Relative to a standalone model,
\begin{equation}
 \Delta L=L(f_{\mathrm{hyb}})-L(f_{\mathrm{base}}), \qquad
 \Delta C=C(f_{\mathrm{hyb}})-C(f_{\mathrm{base}}).
\end{equation}
A hybrid with $\Delta L>0$ and $\Delta C>0$ is dominated in the measured implementation: it is less accurate and more expensive. Runtime is treated as an implementation-level operational measure.

\section{Experimental Design}
\subsection{Data and Target}
We evaluate SPY, QQQ, IWM, AMD, and XOM using 60-minute OHLCV observations obtained from the Alpha Vantage \texttt{TIME\_SERIES\_INTRADAY} endpoint from January 2020 through January 2023. The extraction uses regular trading hours only (\texttt{extended\_hours=false}) and unadjusted intraday prices (\texttt{adjusted=false}). Timestamps are localized to the America/New\_York time zone. Rows with missing timestamps or closing prices are discarded, duplicate timestamps are resolved by keeping the last observation, and records are then sorted chronologically within each asset. The extraction notebook produces 4,649 post-target feature rows for SPY, QQQ, IWM, and XOM and 4,523 for AMD. The downstream modeling table then adds lagged features through five observations together with three- and six-observation rolling summaries and removes rows with incomplete required predictors, yielding 4,644 and 4,518 model-aligned observations, respectively.

The target is a six-observation rolling variance proxy constructed from close-to-close log returns between consecutive observed regular-session bars,
\begin{equation}
 r_t=\log P_t-\log P_{t-1},
\end{equation}
\begin{equation}
 RV_t^{(6)}=\sum_{i=0}^{5} r_{t-i}^2, \qquad y_t=\log RV_t^{(6)}.
\end{equation}
The forecasting task is one-step-ahead prediction of $y_{t+1}$. Returns are computed across consecutive observed bars without resetting at trading-day boundaries; accordingly, the first observed return of a session spans the overnight or weekend interval from the previous session's final bar. The six-term window can therefore cross a session boundary and is interpreted as a six-observation, rather than six-contiguous-clock-hour, variance proxy. We distinguish this construction from conventional high-frequency realized variance, which is typically formed by aggregating finer intraperiod returns over a fixed calendar horizon \cite{andersen2003modeling}.

Adjacent targets share five of six squared-return components. The shared terms induce mechanical short-lag persistence, so the observed dependence reflects both volatility dynamics and the construction of the rolling target.

\subsection{Models and Experimental Roles}
The six models occupy complementary roles in the diagnostic design. AR(5), MA(5), and ARIMA(1,0,1) provide fixed short-memory statistical references \cite{box2015time}. These specifications are pre-specified in the implementation, held constant across assets, and not selected using the reported rolling losses. The primary statistical first stage is an observation-scale HAR-style model,
\begin{equation}
 \hat y_{t+1}=\beta_0+\beta_1 y_t+\beta_2\bar y_t^{(3)}+\beta_3\bar y_t^{(6)},
\end{equation}
where $\bar y_t^{(3)}$ and $\bar y_t^{(6)}$ are trailing three- and six-observation averages. The coefficients are estimated by ordinary least squares within each expanding training window. The six-observation component uses the same observation count as the target-construction window, while the three-observation component provides an intermediate intraday scale; both are fixed by design rather than selected from reported forecast losses. The resulting current/3/6-observation specification is an intraday-bar adaptation of the heterogeneous-horizon HAR idea \cite{corsi2009simple}.

The raw-sequence neural baseline is a univariate single-layer LSTM followed by a linear output layer. Hidden sizes are $\{16,32,64\}$, sequence lengths are $\{5,10,20\}$, and seeds are $\{42,123,999\}$. Training uses Adam with learning rate $10^{-3}$, MSE loss, batch size 64, patience 5, and minimum improvement threshold $10^{-4}$. Early stopping monitors mean training loss; no separate validation subset is used within a refitting block. The initial fit allows up to 25 epochs; subsequent rolling updates use up to five epochs and warm-start from previous neural parameters. No dropout is used. Input sequences are standardized using a scaler fit only on the current training sequences, and the training target is standardized using its current training-block mean and standard deviation before predictions are transformed back to the original scale. At each update, scaling is re-estimated from the expanded training history; neural parameters are then fine-tuned under the updated scaling without an analytical rescaling adjustment to the network weights.

HAR-style+LSTM uses the same LSTM family but replaces the raw volatility sequence with lagged HAR-style residuals. At each refitting origin, HAR-style is estimated only on the currently available training history, and residual-LSTM training sequences are constructed from the in-sample fitted residuals $\tilde e_{t;k}^{\mathrm{fit}}$ within that history. During the subsequent forecast block, a realized pseudo-out-of-sample HAR-style residual $e_{t+1}^{\mathrm{POOS}}$ is appended to residual history only after its corresponding target becomes observed; it can then enter later one-step residual forecasts within the block. The downstream LSTM receives only this residual history, so the experiment measures the predictive content retained after HAR-style residualization without adding the original volatility state or first-stage forecast to the neural input.

\subsection{Rolling Protocol and Model Selection}
All models are evaluated chronologically. The initial estimation window contains the first 60\% of observations. Parameters are subsequently refit or updated every six observations, while the target remains one-step-ahead $y_{t+1}$. Within each six-observation refitting block, forecasts are generated sequentially one step ahead: newly observed target information becomes available to later forecast inputs, while fitted model parameters remain fixed until the next six-observation update. No random shuffling is used. This design follows the principle of rolling-origin evaluation for dependent time series \cite{tashman2000out,hewamalage2023forecast}.

For each neural family and asset, the grid contains nine architecture configurations (three hidden sizes by three sequence lengths), each evaluated with three seeds, for 27 runs. Hidden-size/sequence-length configurations are selected retrospectively by mean rolling MSE across the three seeds, and reported neural results average the selected architecture over those seeds. The same grid is used for Pure LSTM and HAR-style+LSTM to preserve a matched diagnostic comparison.

Hyperparameters are selected retrospectively from the rolling evaluation period, so the reported neural metrics describe selected rolling pseudo-out-of-sample performance. DM comparisons involving neural models are computed for those selected configurations \cite{hewamalage2023forecast}.

\subsection{Computational Environment and Cost Measurement}
The experiments were executed in Google Colab High-RAM with an NVIDIA A100 GPU. Statistical estimation uses \texttt{statsmodels} on CPU, while LSTM training and inference use PyTorch on GPU. End-to-end runtime is measured with wall-clock timing over the complete rolling runner for a model configuration, including repeated fitting or updating and forecast generation. For HAR-style+LSTM, the measured runner includes HAR-style refitting, residual construction, neural updating, and forecast recombination. Total runtime is the primary system-cost measure. Table-level runtime values are means across the five assets after selecting the reported per-asset model configuration; neural configuration-level runtimes are first averaged across seeds. These measurements characterize the evaluated implementation and hardware configuration and should not be interpreted as general algorithmic-complexity rankings. Reported runtimes characterize execution of the selected configuration and exclude the retrospective architecture-search cost.

\section{Results and Discussion}
\subsection{First-Stage Forecasting Validity}
Table~\ref{tab:grouped} reports grouped performance. HAR-style achieves MSE 0.3049, well below AR (0.5785), MA (0.5421), and ARIMA (0.5854), making it the strongest statistical forecaster in the comparison. Pure LSTM records the lowest selected pseudo-out-of-sample MSE, 0.2649.

\begin{table}[H]
\caption{Grouped forecasting results across five assets. Wins denotes the number of assets on which a model attains the lowest selected MSE.}
\label{tab:grouped}
\centering
\scriptsize
\setlength{\tabcolsep}{3.2pt}
\begin{tabular}{lccccc}
\toprule
Model & MSE & MAE & $R^2_{\mathrm{OOS}}$ & QLIKE & Wins\\
\midrule
AR & 0.5785 & 0.4733 & -0.7620 & -6.8708 & 0\\
MA & 0.5421 & 0.4791 & -0.6513 & -6.9024 & 0\\
ARIMA & 0.5854 & 0.4770 & -0.7829 & -6.8701 & 0\\
HAR-style & 0.3049 & 0.3125 & 0.0697 & -7.0875 & 0\\
Pure LSTM & \textbf{0.2649} & \textbf{0.2900} & \textbf{0.1931} & \textbf{-7.1059} & \textbf{5}\\
HAR-style+LSTM & 0.3594 & 0.3609 & -0.0974 & -7.0656 & 0\\
\bottomrule
\end{tabular}
\end{table}

\begin{figure}[H]
\centering
\includegraphics[width=\columnwidth]{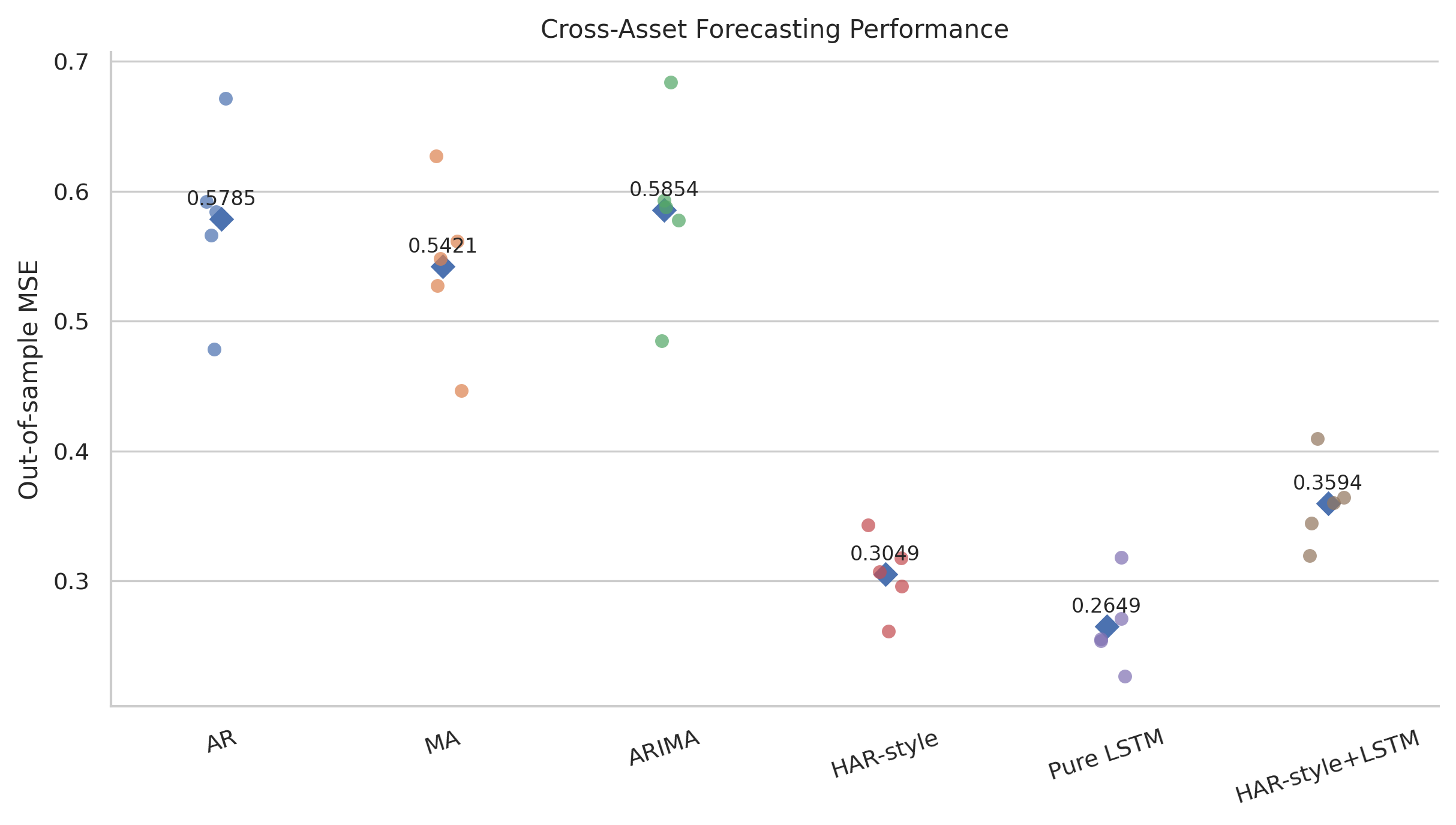}
\caption{Cross-asset MSE for the six-model design. Individual points show assets; diamonds denote cross-asset means.}
\label{fig:performance}
\end{figure}

Figure~\ref{fig:performance} shows the same ordering at the asset level. The generic statistical models occupy the higher-error region, HAR-style lowers error substantially, and Pure LSTM retains the lowest cross-asset selected pseudo-out-of-sample MSE. The improvement is not confined to squared-error loss. HAR-style also reduces MAE to 0.3125 and is the only statistical model with positive benchmark-relative $R^2_{\mathrm{OOS}}$ (0.0697), whereas AR, MA, and ARIMA all fall below the persistence benchmark. Pure LSTM improves further, attaining the lowest MAE and QLIKE and the highest $R^2_{\mathrm{OOS}}$ (0.1931), while recording the lowest selected MSE on all five assets. Taken together, these results establish HAR-style as a credible first-stage forecaster rather than merely a low-MSE statistical reference.

\FloatBarrier
\subsection{Statistical Signal Simplification}
Table~\ref{tab:signal} evaluates both residual objects defined in Sect.~3.1. The fitted residual process before standardization, used to construct residual-LSTM sequences, is strongly compressed: mean within-window variance reduction ranges from 78.97\% to 85.96\% across assets (81.95\% overall), while mean ACF(1) is 0.0089. Diagnostics are computed within each expanding refit window and then averaged within asset, avoiding repeated weighting of older observations. Because HAR-style is estimated by OLS in the same window, this fitted-residual compression describes the handoff process rather than out-of-sample forecast performance. The LSTM subsequently standardizes these sequences, so the variance diagnostic refers to pre-normalization structure. Independently, rolling pseudo-out-of-sample HAR-style errors show 71--77\% variance reduction (74.44\% overall) and mean ACF(1) of $-0.0247$; the magnitude of rolling-error ACF(1) falls by 96.36\% relative to the raw target. The two panels use different aggregation schemes, so each variance-reduction magnitude is interpreted within its own residual object.

\begin{figure}[t]
\centering
\includegraphics[width=\textwidth]{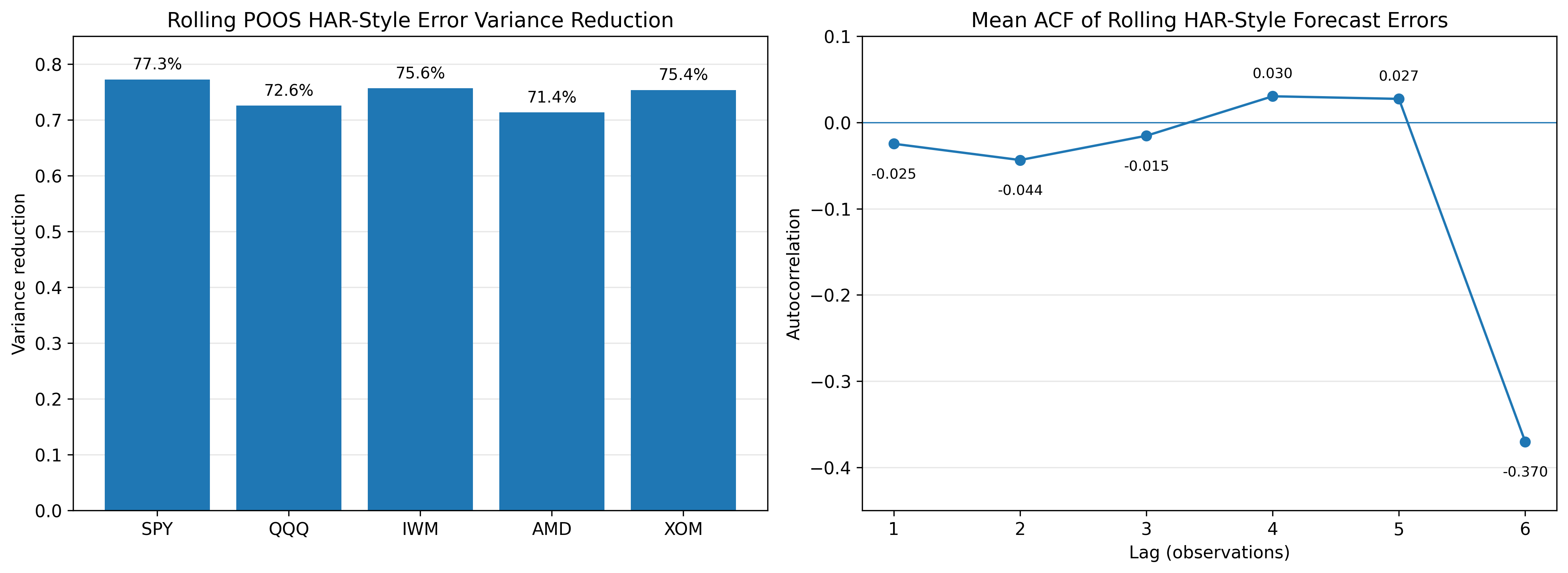}
\caption{Rolling pseudo-out-of-sample HAR-style forecast-error diagnostics. Left: forecast-error variance reduction by asset. Right: cross-asset mean forecast-error autocorrelation at lags 1--6.}
\label{fig:signal}
\end{figure}

Neither residual object is white noise. Mean ACF(6) is $-0.3504$ for fitted training residuals and $-0.3702$ for rolling forecast errors. Figure~\ref{fig:signal} shows the latter across lags 1--6: lags 1--5 are comparatively small, followed by a pronounced negative association at lag 6. The recurring lag-6 pattern may reflect interactions among the six-observation construction, intraday periodicity, and first-stage filtering; we do not attribute it to a single mechanism. Thus, HAR-style materially simplifies the pre-standardization residual process used to construct neural training sequences and independently compresses rolling forecast errors, while structured higher-lag dependence remains.

\begin{table}[t]
\caption{HAR-Style Residual Signal Diagnostics. Training entries are asset-level means across expanding refit windows; POOS entries use sequential rolling forecast errors.}
\label{tab:signal}
\centering
\scriptsize
\setlength{\tabcolsep}{2.3pt}
\begin{tabular}{lcccccc}
\toprule
& \multicolumn{3}{c}{Training fitted residuals} & \multicolumn{3}{c}{Rolling POOS errors}\\
\cmidrule(lr){2-4}\cmidrule(lr){5-7}
Asset & Var. Red. & ACF(1) & ACF(6) & Var. Red. & ACF(1) & ACF(6)\\
\midrule
SPY & 0.8596 & 0.0067 & -0.3303 & 0.7726 & -0.0113 & -0.3584\\
QQQ & 0.8420 & 0.0064 & -0.3424 & 0.7258 & -0.0577 & -0.4070\\
IWM & 0.8144 & 0.0001 & -0.3872 & 0.7565 & -0.0434 & -0.3754\\
AMD & 0.7918 & 0.0123 & -0.3400 & 0.7136 & -0.0275 & -0.3772\\
XOM & 0.7897 & 0.0192 & -0.3521 & 0.7536 & 0.0162 & -0.3329\\
\midrule
Mean & 0.8195 & 0.0089 & -0.3504 & 0.7444 & -0.0247 & -0.3702\\
\bottomrule
\end{tabular}
\end{table}

Whether this statistical simplification is useful for downstream residual prediction is examined next.

\FloatBarrier
\subsection{Forecaster--Preconditioner Asymmetry}
The core comparison is reported in Table~\ref{tab:asset}. HAR-style+LSTM is worse than HAR-style for every asset, with the MSE increase ranging from 0.0372 for SPY to 0.0663 for XOM.

\begin{table}[t]
\caption{Asset-Level Core Diagnostic Comparison}
\label{tab:asset}
\centering
\scriptsize
\setlength{\tabcolsep}{3.0pt}
\begin{tabular}{lcccc}
\toprule
Asset & Pure LSTM & HAR-style & HAR-style+LSTM & $\Delta$MSE\\
\midrule
SPY & 0.2711 & 0.3071 & 0.3443 & +0.0372\\
QQQ & 0.2552 & 0.3175 & 0.3644 & +0.0469\\
IWM & 0.2263 & 0.2612 & 0.3192 & +0.0580\\
AMD & 0.2538 & 0.2959 & 0.3599 & +0.0641\\
XOM & 0.3180 & 0.3430 & 0.4093 & +0.0663\\
\bottomrule
\end{tabular}
\end{table}

\begin{figure}[t]
\centering
\includegraphics[width=\columnwidth]{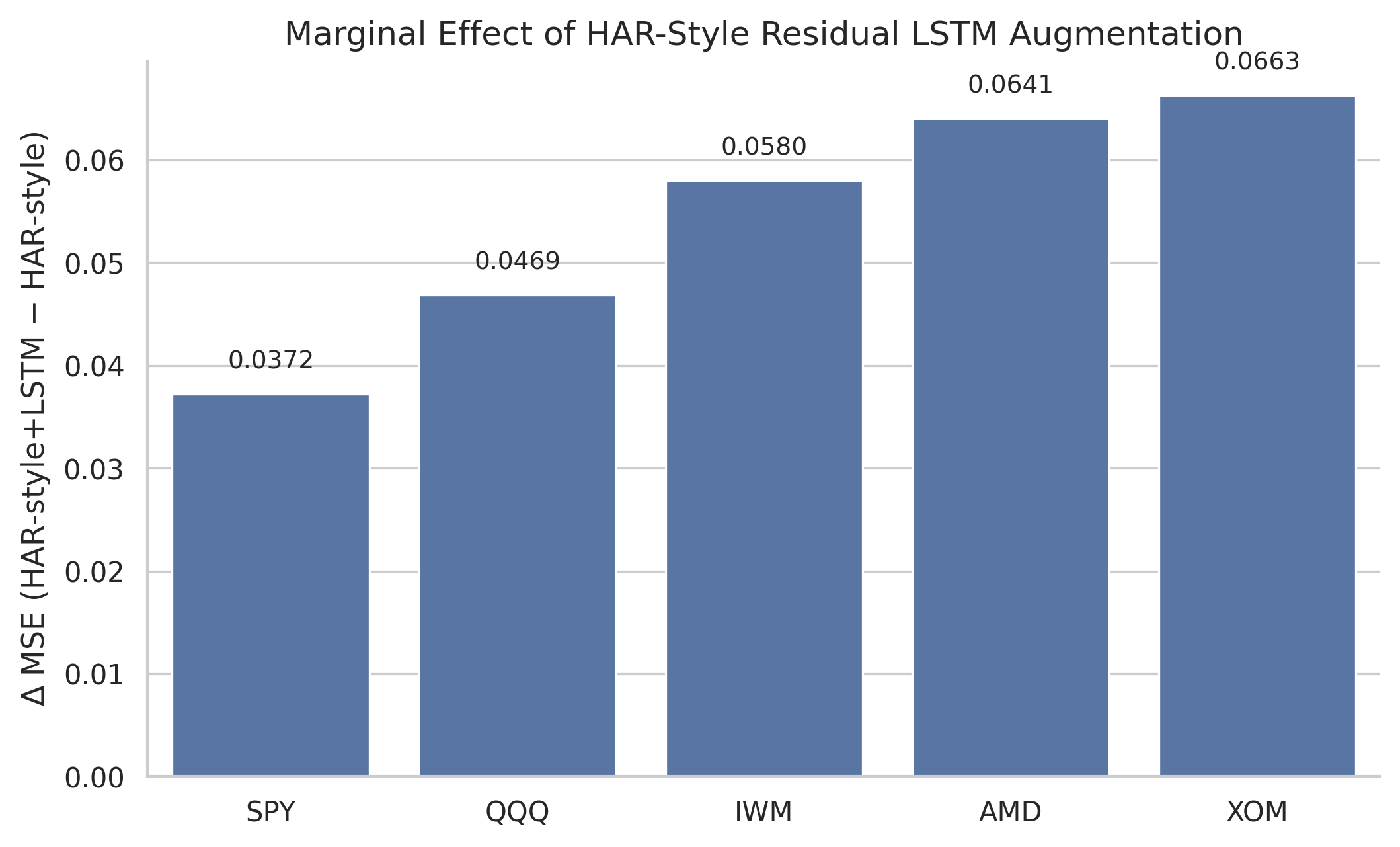}
\caption{Marginal effect of residual LSTM augmentation. $\Delta$MSE is HAR-style+LSTM minus HAR-style; positive values indicate deterioration.}
\label{fig:delta}
\end{figure}

At the grouped level, HAR-style+LSTM raises MSE from 0.3049 to 0.3594, an increase of 0.0545, or 17.9\%. The increase appears on every asset (Fig.~\ref{fig:delta}), not only in the cross-asset mean. Pure LSTM reaches a selected pseudo-out-of-sample MSE of 0.2649, consistent with the raw sequence containing predictive dependence that the selected LSTM can exploit; part of that dependence is mechanically induced by the overlapping target construction described in Sect.~4.1. Relative to HAR-style, the residual hybrid also increases grouped QLIKE by 0.0219 and lowers benchmark-relative $R^2_{\mathrm{OOS}}$ by 0.1671; HAR-style remains above the persistence benchmark on average, whereas HAR-style+LSTM falls below it. In the seed-aggregated asset-level QLIKE comparison, the hybrid--HAR differential is positive on four of five assets and significant on three, so QLIKE is reported as supporting evidence alongside the uniform MSE result. The residual-LSTM search grid includes lookbacks of 5, 10, and 20 observations, placing lag-6 dependence within the available search horizon; longer histories were evaluated but were not selected by the prescribed MSE criterion.

Table~\ref{tab:asymmetry} brings the three layers together. HAR-style combines competitive standalone forecasting with strong compression of the pre-standardization fitted residual process and the rolling forecast errors; after residual-only neural augmentation, MSE and grouped QLIKE rise, benchmark-relative $R^2$ falls, and runtime increases.

\begin{table}[t]
\caption{Forecaster--Preconditioner Asymmetry Summary}
\label{tab:asymmetry}
\centering
\scriptsize
\setlength{\tabcolsep}{3.0pt}
\begin{tabular}{llr}
\toprule
Layer & Diagnostic & Result\\
\midrule
Forecast & HAR-style MSE & 0.3049\\
Signal & Training fitted-residual var. reduction & 81.95\%\\
Signal & Training fitted-residual ACF(1) & 0.0089\\
Signal & Training fitted-residual ACF(6) & -0.3504\\
Signal & Rolling error variance reduction & 74.44\%\\
Signal & Rolling error ACF(1) & -0.0247\\
Signal & Rolling error ACF(6) & -0.3702\\
Forecast & HAR-style+LSTM MSE & 0.3594\\
Forecast & $\Delta$MSE (hybrid - HAR) & +0.0545\\
Forecast & $\Delta$QLIKE (hybrid - HAR) & +0.0219\\
Forecast & $\Delta R^2$ (hybrid - HAR) & -0.1671\\
System & $\Delta$Runtime (s) & +350.25\\
\bottomrule
\end{tabular}
\end{table}

Table~\ref{tab:dm} reports focused asset-level MSE DM comparisons at Newey--West lag 6. For neural contrasts, the three seed-specific loss differentials are averaged at each forecast time and one DM test is then computed per asset; statistical contrasts already have one forecast path per asset. This produces five economically distinct asset-level tests for every comparison. We report the median asset $p$-value and the count with $p<0.05$; all $p$-values below 0.001 are reported as $p<0.001$.

\begin{table}[t]
\caption{Focused Asset-Level Diebold--Mariano Comparisons (MSE Loss; Newey--West Lag 6)}
\label{tab:dm}
\centering
\scriptsize
\setlength{\tabcolsep}{2.0pt}
\begin{tabular}{lcccc}
\toprule
Comparison & Mean Diff. & Median $p$ & $p<0.05$ & $N$\\
\midrule
HAR-style+LSTM vs HAR-style & +0.0545 & $<0.001$ & 5 & 5\\
HAR-style vs AR & -0.2736 & $<0.001$ & 5 & 5\\
HAR-style vs MA & -0.2372 & $<0.001$ & 5 & 5\\
HAR-style vs ARIMA & -0.2805 & $<0.001$ & 5 & 5\\
HAR-style+LSTM vs Pure LSTM & +0.0945 & $<0.001$ & 5 & 5\\
\bottomrule
\end{tabular}
\end{table}

The paired evidence is consistent with the point estimates. After timewise seed aggregation, HAR-style+LSTM versus HAR-style has mean asset-level $\Delta$MSE $=0.0545$, and all five lag-6 asset tests are positive and below 5\%. HAR-style also beats each generic statistical reference in all five asset-level tests, and HAR-style+LSTM remains worse than Pure LSTM on all five assets. Repeating the neural asset-level tests at Newey--West lags 5 and 12 preserves the sign and 5/5 significance pattern, so the result is stable to the HAC bandwidth around the observed lag-6 structure.

One interpretation is that HAR-style residualization changes the location and scale of the predictable structure available to the downstream learner. The pre-standardization fitted residual process used to construct neural training sequences is already much smaller in variance and shows little immediate first-order persistence; the independently generated rolling errors display the same qualitative pattern. Both residual objects nevertheless retain structured dependence at the six-observation lag, and the residual-only LSTM still fails to improve the first-stage forecast. Equation~(5) expresses this interpretation at the signal level; the empirical contrast is between statistical signal transformation and the absence of downstream forecast improvement after the residual sequences are normalized and learned by the LSTM.

\FloatBarrier
\subsection{Accuracy--Efficiency Trade-off}
Table~\ref{tab:efficiency} reports the system-level comparison. HAR-style completes the rolling pipeline in a mean 2.87 seconds with MSE 0.3049. Pure LSTM requires 336.40 seconds and reaches MSE 0.2649. HAR-style+LSTM requires 353.12 seconds and produces MSE 0.3594.

\begin{table}[H]
\caption{Accuracy--Efficiency Summary}
\label{tab:efficiency}
\centering
\scriptsize
\setlength{\tabcolsep}{5.0pt}
\begin{tabular}{lcc}
\toprule
Model & End-to-End Runtime (s) & MSE\\
\midrule
AR & 221.39 & 0.5785\\
MA & 471.51 & 0.5421\\
ARIMA & 159.14 & 0.5854\\
HAR-style & 2.87 & 0.3049\\
Pure LSTM & 336.40 & \textbf{0.2649}\\
HAR-style+LSTM & 353.12 & 0.3594\\
\bottomrule
\end{tabular}
\end{table}

\begin{figure}[H]
\centering
\includegraphics[width=\columnwidth]{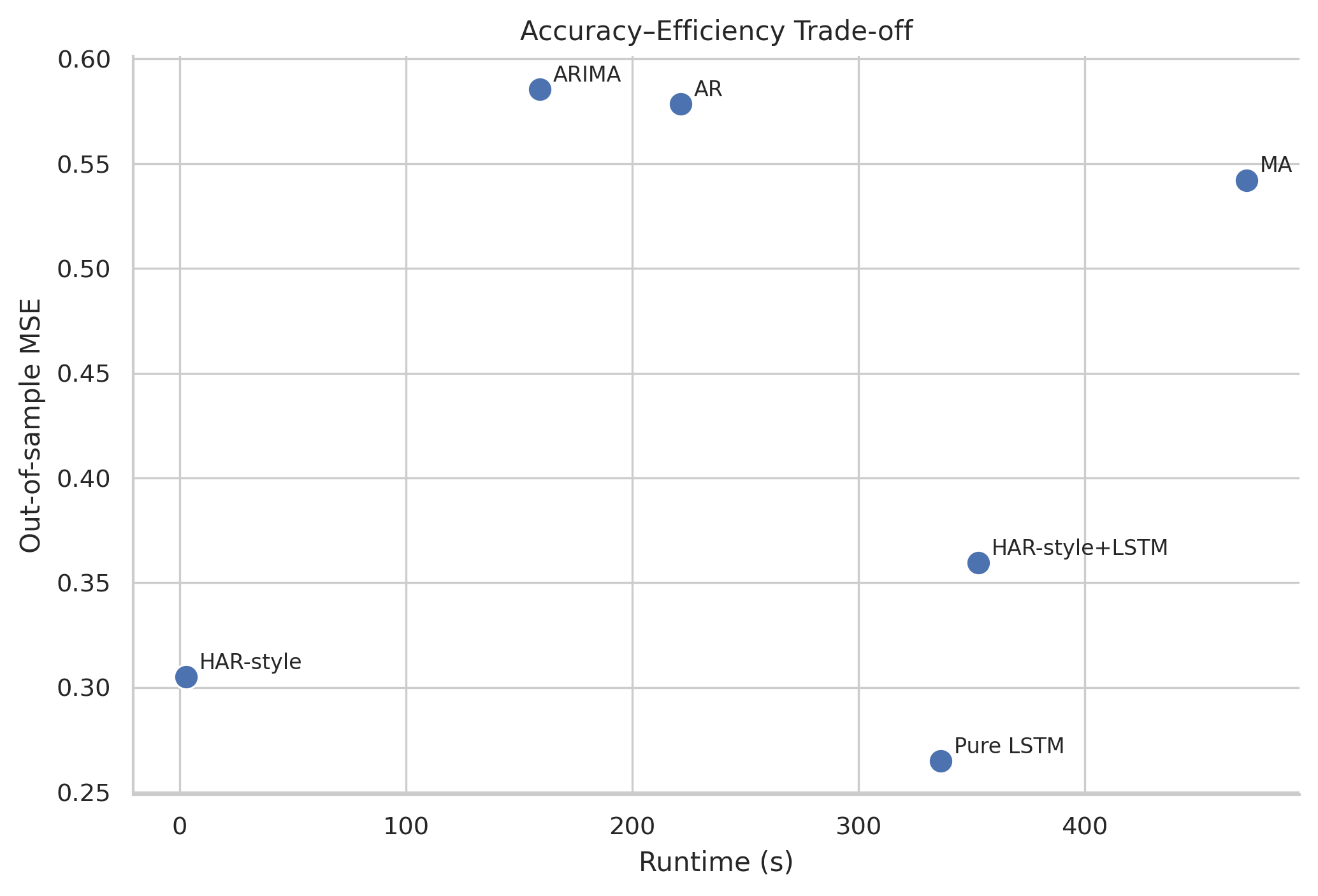}
\caption{Accuracy--efficiency trade-off across the six models. Lower-left is preferred. HAR-style+LSTM is less accurate than both HAR-style and Pure LSTM while retaining neural-scale runtime.}
\label{fig:efficiency}
\end{figure}

Figure~\ref{fig:efficiency} makes the trade-off explicit. Relative to HAR-style, the residual hybrid has both $\Delta L>0$ and $\Delta C>0$; relative to Pure LSTM, it has similar end-to-end runtime but worse predictive performance. The additional computation therefore does not purchase a complementary forecasting advantage. HAR-style represents the low-cost statistical solution, whereas Pure LSTM represents the accuracy-oriented neural solution. HAR-style+LSTM retains neural-scale training cost but matches neither the accuracy of the raw-sequence LSTM nor the efficiency of HAR-style, placing it outside both favorable regions of the evaluated design space.

\FloatBarrier
\section{Interpretation and Implications}
The empirical distinction matters because hybrid design choices are often made from first-stage fit alone. On that criterion, HAR-style residualization looks attractive: it produces the strongest statistical forecast, strongly compresses the pre-standardization fitted residual process used to construct neural training sequences, and yields similarly compressed rolling forecast errors. The downstream comparison changes the picture. Once the neural learner receives the standardized residual history, the transformed representation supports less accurate forecasts than either HAR-style alone or Pure LSTM on the raw sequence.

One explanation is that residualization changes which temporal structures remain accessible to the neural stage. Equation~(5) gives this intuition a signal-level form: residual variance can fall even as the predictable share of the remaining variation becomes less useful to the second stage. The similar negative lag-6 dependence observed in both fitted training residuals and rolling forecast errors also shows that statistical simplification should not be equated with whitening. The present diagnostics identify the pipeline-level contrast without uniquely decomposing the information removed or retained by the first stage.

The residual-only design keeps the information handoff transparent. A context-aware hybrid that also receives the original volatility state, the first-stage forecast, or cross-asset information would answer a different question: whether added context can recover value after residualization. The same distinction can therefore be tested in richer multivariate or nonlinear architectures without changing the diagnostic logic.

The empirical design spans five U.S. assets over 2020--2023 with the overlapping six-observation target in Sect.~4.1. The neural experiments use a single-layer univariate LSTM family, keeping the information handoff directly observable and defining the architecture class addressed by the evidence. Configurations are selected retrospectively from the rolling evaluation period; results are therefore reported as selected pseudo-out-of-sample performance, with DM tests computed for those selected configurations. HAC-bandwidth checks at lags 5, 6, and 12 verify the serial-dependence robustness of the MSE comparisons. Training sequences use contemporaneously fitted HAR residuals, while deployment history incorporates realized one-step forecast errors, making the fitted-to-realized transition an explicit component of the evaluated pipeline. Prices are unadjusted, so overnight returns retain both market movements and any distribution- or corporate-action-related price changes. The same diagnostic design can be extended to separate chronological validation, broader asset panels, adjusted or session-aware returns, and additional sequence architectures \cite{hewamalage2023forecast}.

For hybrid financial forecasting, the key design question sits at the information handoff between stages. A strong first-stage model and a capable neural learner do not guarantee that the intermediate representation is useful. Evaluating the transformed signal together with downstream loss and system cost makes that handoff explicit.

\section{Conclusion}

Statistical residual simplification does not necessarily create a more useful neural forecasting target. HAR-style is the strongest standalone statistical forecaster and materially simplifies both the fitted residual process and rolling forecast errors: pre-standardization fitted-residual variance falls by approximately 82\%, rolling forecast-error variance falls by approximately 74\%, and lag-1 autocorrelation is near zero in both. Yet residual-only LSTM augmentation raises mean MSE from 0.3049 to 0.3594, a 17.9\% deterioration, with higher error on all five assets. Seed-aggregated asset-level DM tests confirm the deterioration across all five assets, and the result remains significant under Newey--West lags 5, 6, and 12. Pure LSTM achieves the lowest selected pseudo-out-of-sample MSE, 0.2649.

These results reveal a forecaster--preconditioner asymmetry. A first-stage model can forecast well and substantially simplify the residual process without producing a representation that benefits downstream neural correction. Forecasting quality and preconditioning quality are therefore distinct properties. In this setting, HAR-style succeeds as a forecaster but does not create a residual representation that supports beneficial LSTM augmentation.

The system-level evidence reinforces the same conclusion. HAR-style provides the low-cost statistical solution, Pure LSTM provides the strongest predictive solution, and HAR-style+LSTM combines neural-scale runtime with higher forecast error. More broadly, hybrid financial forecasting should be evaluated at the information handoff between stages: the value of a hybrid is established by downstream predictive performance, not inferred from the strength of its individual components or from residual simplification alone.

\bibliographystyle{splncs04}
\bibliography{refs}

@article{andersen2003modeling,
  author={T. G. Andersen and T. Bollerslev and F. X. Diebold and P. Labys},
  title={Modeling and forecasting realized volatility},
  journal={Econometrica},
  volume={71},
  number={2},
  pages={579--625},
  year={2003},
  note={doi: 10.1111/1468-0262.00418}
}

@article{bai2018empirical,
  author={S. Bai and J. Z. Kolter and V. Koltun},
  title={An empirical evaluation of generic convolutional and recurrent networks for sequence modeling},
  journal={arXiv:1803.01271},
  year={2018}
}

@article{bollerslev1986generalized,
  author={T. Bollerslev},
  title={Generalized autoregressive conditional heteroskedasticity},
  journal={Journal of Econometrics},
  volume={31},
  number={3},
  pages={307--327},
  year={1986},
  note={doi: 10.1016/0304-4076(86)90063-1}
}

@book{box2015time,
  author={G. E. P. Box and G. M. Jenkins and G. C. Reinsel and G. M. Ljung},
  title={Time Series Analysis: Forecasting and Control},
  edition={5th},
  address={Hoboken, NJ, USA},
  publisher={Wiley},
  year={2015}
}

@article{bucci2020realized,
  author={A. Bucci},
  title={Realized volatility forecasting with neural networks},
  journal={Journal of Financial Econometrics},
  volume={18},
  number={3},
  pages={502--531},
  year={2020},
  note={doi: 10.1093/jjfinec/nbz014}
}

@article{christensen2023machine,
  author={K. Christensen and M. Siggaard and B. Veliyev},
  title={A machine learning approach to volatility forecasting},
  journal={Journal of Financial Econometrics},
  volume={21},
  number={5},
  pages={1680--1727},
  year={2023},
  note={doi: 10.1093/jjfinec/nbac020}
}

@article{corsi2009simple,
  author={F. Corsi},
  title={A simple approximate long-memory model of realized volatility},
  journal={Journal of Financial Econometrics},
  volume={7},
  number={2},
  pages={174--196},
  year={2009},
  note={doi: 10.1093/jjfinec/nbp001}
}

@article{diebold1995comparing,
  author={F. X. Diebold and R. S. Mariano},
  title={Comparing predictive accuracy},
  journal={Journal of Business \& Economic Statistics},
  volume={13},
  number={3},
  pages={253--263},
  year={1995},
  note={doi: 10.1080/07350015.1995.10524599}
}

@article{engle1982autoregressive,
  author={R. F. Engle},
  title={Autoregressive conditional heteroscedasticity with estimates of the variance of {United Kingdom} inflation},
  journal={Econometrica},
  volume={50},
  number={4},
  pages={987--1007},
  year={1982},
  note={doi: 10.2307/1912773}
}

@article{hansen2006realized,
  author={P. R. Hansen and A. Lunde},
  title={Realized variance and market microstructure noise},
  journal={Journal of Business \& Economic Statistics},
  volume={24},
  number={2},
  pages={127--161},
  year={2006},
  note={doi: 10.1198/073500106000000071}
}

@article{hochreiter1997long,
  author={S. Hochreiter and J. Schmidhuber},
  title={Long short-term memory},
  journal={Neural Computation},
  volume={9},
  number={8},
  pages={1735--1780},
  year={1997},
  note={doi: 10.1162/neco.1997.9.8.1735}
}

@article{khashei2011novel,
  author={M. Khashei and M. Bijari},
  title={A novel hybridization of artificial neural networks and {ARIMA} models for time series forecasting},
  journal={Applied Soft Computing},
  volume={11},
  number={2},
  pages={2664--2675},
  year={2011},
  note={doi: 10.1016/j.asoc.2010.10.015}
}

@article{kim2018forecasting,
  author={H. Y. Kim and C. H. Won},
  title={Forecasting the volatility of stock price index: A hybrid model integrating {LSTM} with multiple {GARCH}-type models},
  journal={Expert Systems with Applications},
  volume={103},
  pages={25--37},
  year={2018},
  note={doi: 10.1016/j.eswa.2018.03.002}
}

@article{lim2021timeseries,
  author={B. Lim and S. Zohren},
  title={Time-series forecasting with deep learning: A survey},
  journal={Philosophical Transactions of the Royal Society A},
  volume={379},
  number={2194},
  year={2021},
  note={Art. no. 20200209, doi: 10.1098/rsta.2020.0209}
}

@article{newey1987simple,
  author={W. K. Newey and K. D. West},
  title={A simple, positive semi-definite, heteroskedasticity and autocorrelation consistent covariance matrix},
  journal={Econometrica},
  volume={55},
  number={3},
  pages={703--708},
  year={1987},
  note={doi: 10.2307/1913610}
}

@inproceedings{nie2023time,
  author={Y. Nie and N. H. Nguyen and P. Sinthong and J. Kalagnanam},
  title={A time series is worth 64 words: Long-term forecasting with transformers},
  booktitle={Proc. Int. Conf. Learning Representations ({ICLR})},
  year={2023}
}

@inproceedings{oreshkin2020nbeats,
  author={B. N. Oreshkin and D. Carpov and N. Chapados and Y. Bengio},
  title={{N-BEATS}: Neural basis expansion analysis for interpretable time series forecasting},
  booktitle={Proc. Int. Conf. Learning Representations ({ICLR})},
  year={2020}
}

@article{patton2011volatility,
  author={A. J. Patton},
  title={Volatility forecast comparison using imperfect volatility proxies},
  journal={Journal of Econometrics},
  volume={160},
  number={1},
  pages={246--256},
  year={2011},
  note={doi: 10.1016/j.jeconom.2010.03.034}
}

@article{sezer2020financial,
  author={O. B. Sezer and M. U. Gudelek and A. M. Ozbayoglu},
  title={Financial time series forecasting with deep learning: A systematic literature review: 2005--2019},
  journal={Applied Soft Computing},
  volume={90},
  year={2020},
  note={Art. no. 106181, doi: 10.1016/j.asoc.2020.106181}
}

@article{zhang2003time,
  author={G. P. Zhang},
  title={Time series forecasting using a hybrid {ARIMA} and neural network model},
  journal={Neurocomputing},
  volume={50},
  pages={159--175},
  year={2003},
  note={doi: 10.1016/S0925-2312(01)00702-0}
}

@article{tashman2000out,
  author={L. J. Tashman},
  title={Out-of-sample tests of forecasting accuracy: an analysis and review},
  journal={International Journal of Forecasting},
  volume={16},
  number={4},
  pages={437--450},
  year={2000},
  note={doi: 10.1016/S0169-2070(00)00065-0}
}

@article{hewamalage2023forecast,
  author={H. Hewamalage and K. Ackermann and C. Bergmeir},
  title={Forecast evaluation for data scientists: common pitfalls and best practices},
  journal={Data Mining and Knowledge Discovery},
  volume={37},
  pages={788--832},
  year={2023},
  note={doi: 10.1007/s10618-022-00894-5}
}

@article{branco2024forecasting,
  author={Rafael R. Branco and Alexandre Rubesam and Mauricio Zevallos},
  title={Forecasting realized volatility: Does anything beat linear models?},
  journal={Journal of Empirical Finance},
  volume={78},
  pages={101524},
  year={2024},
  doi={10.1016/j.jempfin.2024.101524}
}

@article{luo2024forecasting,
  author={Jiawen Luo and Tony Klein and Thomas Walther and Qiang Ji},
  title={Forecasting realized volatility of crude oil futures prices based on machine learning},
  journal={Journal of Forecasting},
  volume={43},
  number={5},
  pages={1422--1446},
  year={2024},
  doi={10.1002/for.3077}
}

@article{gunnarsson2024prediction,
  author={Elias S. Gunnarsson and Hakon R. Isern and Aristidis Kaloudis and Morten Risstad and Benjamin Vigdel and Sjur Westgaard},
  title={Prediction of realized volatility and implied volatility indices using AI and machine learning: A review},
  journal={International Review of Financial Analysis},
  volume={93},
  pages={103221},
  year={2024},
  doi={10.1016/j.irfa.2024.103221}
}

\end{document}